# Surrogate Model For Field Optimization Using Beta-VAE Based Regression

Ajitabh Kumar

Visage Technology

## Abstract

Oilfield development related decisions are made using reservoir simulation-based optimization study in which different production scenarios and well controls are compared. Such simulations are computationally expensive and so surrogate models are used to accelerate studies. Deep learning has been used in past to generate surrogates, but such models often fail to quantify prediction uncertainty and are not interpretable. In this work, beta-VAE based regression is proposed to generate simulation surrogates for use in optimization workflow. beta-VAE enables interpretable, factorized representation of decision variables in latent space, which is then further used for regression. Probabilistic dense layers are used to quantify prediction uncertainty and enable approximate Bayesian inference.

A single optimization run requires thousands of compute-expensive simulation runs to calculate a real valued objective function such as net present value or weighted cumulative fluid production. Surrogate models may be used in place of full-physics simulation to accelerate optimization studies or quickly analyze different scenarios. With this goal in mind, beta-VAE based regression is used in which factorized representation in latent space makes the model interpretable while probabilistic layers determine the prediction uncertainty. Latent factors are disentangled by increasing value of beta which enhances interpretability at the cost of larger reconstruction error. Probabilistic dense layers assist in modeling high-dimensional decision variables and ascertaining uncertainty. Latent representation is further used as input in regression, and both VAE as well as regression networks are trained jointly. This ensures that encoder finds meaningful latent representation that is useful for downstream supervised learning task.

Olympus benchmark is used to validate our approach with test case comprising of 72 well control and 18 well placement related decision variables. Two separate objective functions namely weighted cumulative fluid production and net present value of cash flow are used. As beta value is increased from value of one which corresponds to original VAE, latent variable representation first improves and then diffuses for even higher value. Thus, proper selection of beta value could make the factors independent and hence interpretable. Probabilistic dense layers enable modeling of complex, high-dimensional, and trainable priors as well as quantification of prediction uncertainty. Both predicted objective function value and associated uncertainty can be used to decide whether predictions are acceptable or full simulation should be made for a case. Thus, optimization study can be accelerated without compromising the quality of results.

Surrogate model developed using beta-VAE based regression finds interpretable and relevant latent representation. A reasonable value of beta ensures a good balance between factor disentanglement and reconstruction. Probabilistic dense layer helps in quantifying predicted uncertainty for objective function, which is then used to decide whether full-physics simulation is required for a case.

## Introduction

Many business decisions in oil and gas industry utilize reservoir simulation-based oilfield production forecast. Hundreds, or even thousands, of full-physics simulation runs are made to optimize across production scenarios and to quantify uncertainty. This makes the whole process compute- and labor-intensive. While automated studies cut down on the labor cost of such studies, compute- and time-costs still remain high.

Simulation study uses a reservoir model, which is a simplified computational representation of oilfield based on data from geological analogs, seismic, well logs, and other dynamic behavior. Physical infrastructure including wells are coded into the model, and simulation-based optimization workflow tunes parameters including well location and control parameters to maximize the total production output. Recently, machine learning and cloud computing have been used with reservoir simulation to bring down compute- and time-cost respectively. In this work, β-VAE based regression is used to generate uncertainty-aware surrogate models which give prediction for the simulated response. Based on the predicted value and its uncertainty, a need for computationally expensive full-physics simulation may be obviated for a given case.

Reservoir simulation studies are carried out to make various business decisions spanning different operational time-scales. Short-time scale decisions are related to well production control, while medium time-scale decisions are related to infill well or completion designs among others. Long term decisions include field development strategies, lifecycle planning and asset management. Previous works have looked into ways of accelerating such studies and fall mainly into two categories: reduced order modeling (ROM) (He et al., 2015) and polynomial interpolation techniques (Valladao et al., 2013). Recently deep neural network has been used to develop surrogate model for a SPE10 benchmark using LSTM with encoder-decoder architecture (Navrátil et al., 2019).

Deep neural networks typically use thousands of training samples and resulting models are not interpretable. This is in contrast with reservoir simulation modeling where physical rules are coded into simulators, and thus results are interpretable. Rule (or physics) based programming is the basis of such computational tools and builds the foundation of symbolic-AI based expert systems. But such tools are limited by the understanding of systems and suffer from knowledge acquisition problem. Modern machine learning based models, on the other hand, use large amounts of unstructured data and yield satisfactory results in various applications. But such DNN models do not generalize well to unseen situation, i.e. out-of-distribution samples, which is in contrast with human intelligence which generalizes beyond one's experiences. It is argued that combinatorial generalization is essential for AI to match human-like abilities. Human ability to generalize greatly depends on representing complex systems as composition of entities and their interaction. Having a data representation suitable for a given machine learning task can significantly improve the learning success and robustness of the model. Recently, beta-VAE has been proposed to learn interpretable factorized representation of independent data generative factors of system in an unsupervised manner. Beta-VAE is basically a modification of variational autoencoder (VAE) framework which yields disentangled representation, and includes an additional hyperparameter $β$ (Higgins et al., 2017). Other such disentangled representation learning efforts include AnnealedVAE, FactorVAE, β-TCVAE and DIP-VAE (Burgess et al., 2018; Kim et al., 2019; Chen et al., 2019; Kumar et al., 2018). Recently, performance of such models have also been studied for generating correlated latent factors (Trauble et al., 2021).

Initial applications of deep network approach have been for non-critical applications and hence prediction uncertainty quantification was not emphasized. As these models are now applied in critical industries like health, law, and self-driving, it becomes important to capture prediction uncertainty (Dusenberry et al., 2020; Heo at al., 2018). It would be desirable to use alternate modeling approach or even seek human intervention when the prediction uncertainty of a given DNN model is high. For a capital-intensive industry like oil & gas, incorrect or even sub-optimal decision can be quite costly. Recent works on DNN prediction uncertainty show that deep networks with dropout sampling (Gal et al., 2016) and batch normalization (Teye et al., 2018) can be considered as Bayesian neural network. Another approach for capturing uncertainty is to use probabilistic neural layer with distributions for network parameters (Dillon et al., 2017). Deep ensembles with models trained using different random initialization have also proved to be quite efficient for uncertainty assessment (Lakshminarayanan et al., 2017; Malinin, 2019). DNN models have two kinds of prediction uncertainty: aleatoric (or data) uncertainty which captures measurement noise that cannot be reduced, and epistemic (or model) uncertainty which is systematic in modeling approach used and can be reduced using a larger dataset (Ovadia et al., 2019). Uncertainty quantification of DNN model prediction is required for use in critical systems.

In this work, surrogate models are developed using DNN based regression model which outputs real-valued parameter for a given production scenario. β-VAE has been used in previous studies to generate factorized latent representation, while regression model is used to predict desired output parameter for given set of inputs. In this work, both β-VAE and regression model are trained simultaneously using a joint objective function (Zhao et al., 2019; Cai et al., 2020). Thus, latent representation not only tries to reproduce input features as in standard VAE, but also aims to capture the correct latent embedding for further use in regression. The main contributions of this work are:

- Uncertainty aware surrogate models are developed which may be used to accelerate reservoir simulation workflow.
- β-VAE based regression model is trained with a combined optimization objective which reduces gap between unsupervised and supervised learning by finding meaningful, factorized latent representation of input features.
- Model is extended to learn correlated factors which could help in incorporating causality.
- Prediction uncertainty is captured both using monte-carlo dropout and probabilistic neural layers.

## Approach

In this section, both the reservoir modeling and deep neural network modeling approaches used in this study are discussed. Olympus reservoir model is used to conduct field optimization study, and data harvested from the study are used to develop DNN model.

### Reservoir Simulation Model

Olympus benchmark case is a synthetic simulation model based on a typical North Sea offshore oilfield (Fonseca et al., 2020). Field is a 9km by 3km by 50m in size with minor faults and limited aquifer support. In addition to the boundary fault, six smaller, internal faults are present in the reservoir. The reservoir is 50 m thick and consists of two zones that are separated by an impermeable shale layer. The upper reservoir zone contains fluvial channel sands embedded in floodplain shales, while the lower reservoir zone consists of alternating layers of coarse, medium, and fine sands. Simulation model has 11 producer and 7 injector wells, all on bottom hole pressure constraint (Figure 1).

Dataset for this work is harvested from a separate Olympus field optimization study of a redevelopment exercise. Optimization problem includes identifying location of three production wells for field redevelopment exercise, while the other fifteen wells are treated as existing wells. Well location is defined using two 3-D coordinate sets indicating heel and toe for any well, thus having total six variables for a well. Simultaneously, bottom-hole pressure (BHP) value for production control is also optimized for all the 18 wells at four predetermined times during the 20 year simulation period. Thus, joint optimization problem has 18 placement related and 72 control related variables, giving a total 90 of real-valued decision variables.

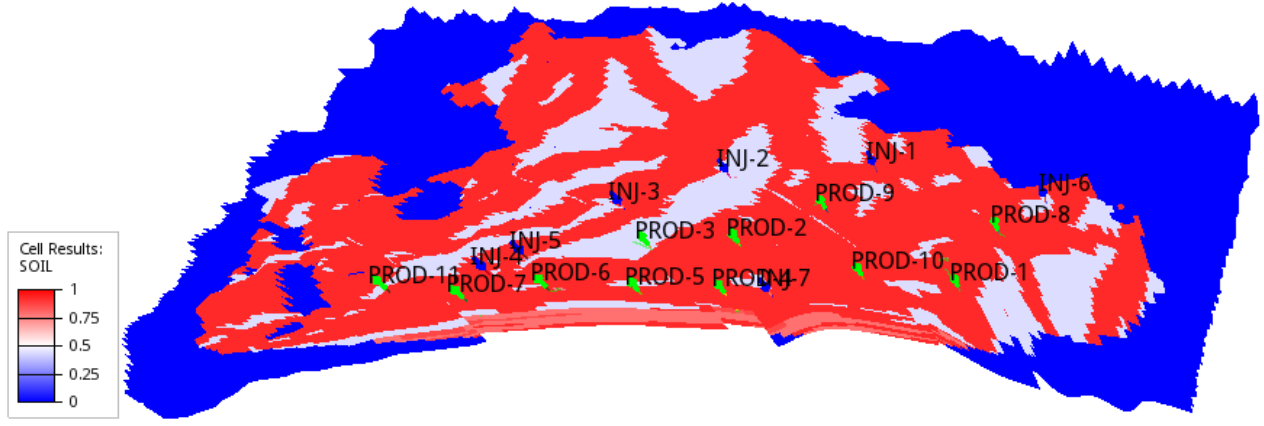

**Figure 1:** Olympus simulation model overview: initial oil saturation and well locations

Two different objective functions are used in optimization study, first being the weighted sum of cumulative fluid production/injection while second being the net present value of discounted cash-flow (Baumann et al., 2020; Tanaka et al., 2018; Udy et al., 2017). This is done to have a rigorous analysis by encoding varying degree of non-linearity among the two objective functions. Weighted sum of cumulative fluid (WCF) is defined as:

$$WCF = Q_{op} - 0.1 \times (Q_{wp} + Q_{wi})$$

where $Q_{op}$ is total oil produced, $Q_{wp}$ is total water produced, and $Q_{wi}$ is total water injected. Net present value (NPV) of discounted cash-flow incorporates additional degree of non-linearity and is defined as:

$$NPV = \sum_{f=1}^{N_f} \sum_{\tau=1}^{N_\tau} \frac{Q_{f,\tau} C_f}{(1+d_f)^{\tau-1}} + \sum_{i=1}^{N_w} D_i$$

where $N_f$ is the number of fluid component, $N_\tau$ is number of time intervals, $Q_{f,\tau}$ is the volume component produced/injected in time interval τ, $C_f$ is the price/cost while $d_f$ is discount factor for component $f$, $N_w$ is number of wells drilled, and $D_i$ is drilling cost for well $i$.

Dataset for deep learning has been harvested from this automated field optimization study of Olympus model. Training dataset is sparsely populated for low objective values in both WCF and NPV cases, as well as for high values in WCF case.

## Neural Network Model

Goal of DNN modeling in this study is to use well placement and control parameters as input feature, and predict objective function value at the end of simulation run as a single, real-valued output parameter. β-VAE is used to generate factorized latent representation of input features, which is then used in regression. Both VAE and regression models are trained simultaneously using a joint objective function so that latent representation is relevant to the regression task. This approach reduces the gap between unsupervised and supervised learning, and should improve learning success and robustness of DNN model (Locatello et al., 2019).

VAE model also works as dimensionality reduction (or data compression) tool. Oilfield optimization problem has 90 decision variables, and interpreting such large dimensional input feature is difficult. Also, population-based optimization algorithms such as particle swarm optimization and genetic algorithms require larger number of iterations for a high dimensional problem. In continuous domains, the volume of search space grows exponentially with dimensions, and is known as the curse of dimensionality. This phenomenon causes a population size that is capable of adequate search space coverage in low dimensional search spaces to degrade into sparse coverage in high dimensional search spaces (Chen et al., 2015). VAE model enables representation of high dimensional input feature into low-dimensional latent features, which in turn should enhance interpretability (Ghojogh et al., 2021; Rolinek et al., 2019; Van Der Maaten et al., 2009). Latent variable models also enable us to code structural relationship by injecting inductive bias and prior knowledge (Locatello et al., 2019; Ridgeway, 2016; Tschannen et al., 2018; Zietlow et al., 2021).

Aleatoric or data uncertainty is present in the simulated dataset because of solver residual and numerical truncation errors. Epistemic or model uncertainty, on the other hand, is present because of uncertainty around true function underlying the observed process. While model uncertainty can be reduced using a larger dataset, data uncertainty cannot be reduced (Malinin, 2019). DNN model uncertainty is captured using monte-carlo dropout for deterministic dense hidden layers, and then using probabilistic dense layers.

Variational autoencoder network with three dense hidden layers each in encoder and decoder is used, and similarly regressor network also includes three hidden layers. This was done to get the desired latent representation of features to be used in conjugation with the regression model. A joint objective function was used as the sum of KL regularization term and mean squared errors for VAE and regression. Hyperparameter tuning exercise confirmed that batch normalization and dropout are required for regularization and stable learning. Hence, batch normalization, leaky relu activation, and dropout, in that order, are used after dense layer in all the models. The mean-field latent representation of VAE assumes independent Gaussian factors, which might not be a valid assumption for this highly-nonlinear system with wells interacting within reservoir across the given time span. So, latent parameters in the probabilistic model are considered as correlated multi-variate normal so as to capture factor causality (Trauble et al., 2021).

## Algorithm and Architecture

In this section, key algorithmic concepts used in the study are discussed with focus on factorized latent representation and uncertainty assessment using Bayesian approach. Different neural network architectures used are described next along with their implementation details.

### Algorithm

Bayesian neural network works by first assigning a prior to the network parameters and then training model to find posterior distribution. Exact posterior is not available because of non-linearity and non-conjugacy, and hence approximate Bayesian inference methods are used (Gal et al., 2016).

Variational inference approximates the true intractable posterior by finding a surrogate distribution that minimizes KL divergence to the posterior (Kingma et al., 2013). Unlike MCMC which is based on sampling, variational approach solves an optimization problem. Posterior is generally approximated using independent mean-field or factorized distribution. Use of correlated factors with multi-variate normal distribution has been proposed to better capture the underlying system component interactions.

$\beta$-VAE is a modification of VAE framework and introduces an adjustable hyperparameter $\beta$. This parameter balances the latent channel capacity and factor independence with reconstruction accuracy. A value of 1 for this parameter relates to standard VAE, while larger beta value pushes the model to learn a more efficient latent representation with disentangled factors (Higgins et al., 2017; Higgins et al., 2018; Burgess et al, 2018). This extra pressure from $\beta$ creates a trade-off between reconstruction accuracy and quality of disentanglement within the latent representation. $\beta$ value is tuned to find the right balance between information preservation (using reconstruction cost) and latent channel capacity restriction (using $\beta>1$). A disentangled representation can be understood as one latent factor being sensitive to one generating factor, while being relatively invariant to other factors. Advantages of disentangled representation include: (i) invariance i.e. representation is invariant to irrelevant factors, (ii) tranferability i.e. more suitable for transfer learning key generative factors, (iii) interpretability i.e. assign meanings to factors, (iv) data efficiency, and (v) compositionality i.e. ability to ignore certain factors or use novel factor recombinations for better generalization.

Monte-carlo dropout is a combination of two algorithms, dropout in which each neuron is independently zeroed out at every forward pass with a given probability, and monte-carlo in which posterior is estimated by sampling (Gal et al., 2016). Trained model is run multiple times to obtain a distribution for a specific input.

Probabilistic dense layers replace deterministic weights of dense layers with probability distributions which enable model to capture uncertainties. Recently deep learning probabilistic frameworks have been developed, such as Tensorflow Probability, which are scalable and flexible enough for real world use (Dillon et al., 2017).

**Architecture**

β-VAE based regression is used in this study with all encoder, decoder, and regressor having three hidden layers each. Batch normalization, leaky relu activation and dropout layer are used following each dense layer for regularization and stable learning. Mean squared error for predicted output is used as objective function while ADAM algorithm is used for optimization. A combined loss function is defined as the sum of mean squared input feature reconstruction error along with KL-regularization loss for the latent space (VAE error), and mean squared output prediction error (regression error) as shown in the following equation (Cai et al., 2020; Zhao et al., 2019).

$$Loss = MSE(x, \hat{x}) + \beta \cdot D_{KL}(q_\phi(z) \| p_\theta(z)) + \gamma \cdot MSE(y, \hat{y})$$

where $x$ is the input feature, $y$ is the target variable, $\hat{x}$ is the reconstructed feature, $\hat{y}$ is the predicted output, $z$ is the latent representation, $p$ is the normal prior, $q$ is the approximate posterior, $\phi$ is the variational parameter, $\theta$ is the model parameter, $D_{KL}$ is the KL-regularization loss, $MSE(x,\hat{x})$ is the mean-squared reconstruction error, and $MSE(y,\hat{y})$ is the mean-squared prediction error. $\beta$ parameter scales the KL loss, while $\gamma$ parameter scales the squared regression error. KL-regularization loss is further calculated as (Kingma et al., 2013):

$$D_{KL}(q_\phi(z) \| p_\theta(z)) = -0.5 \sum_{j=1}^{J} \left(1 + \log(\sigma_j^2) - \mu_j^2 - \sigma_j^2\right)$$

where $\mu_j$ and $\sigma_j$ represent mean and variance of Gaussian latent embedding for its vector element $j$. Figure 2 shows the schematic for β-VAE based regression framework. Once model has been trained, dropout layers enable prediction uncertainty quantification.

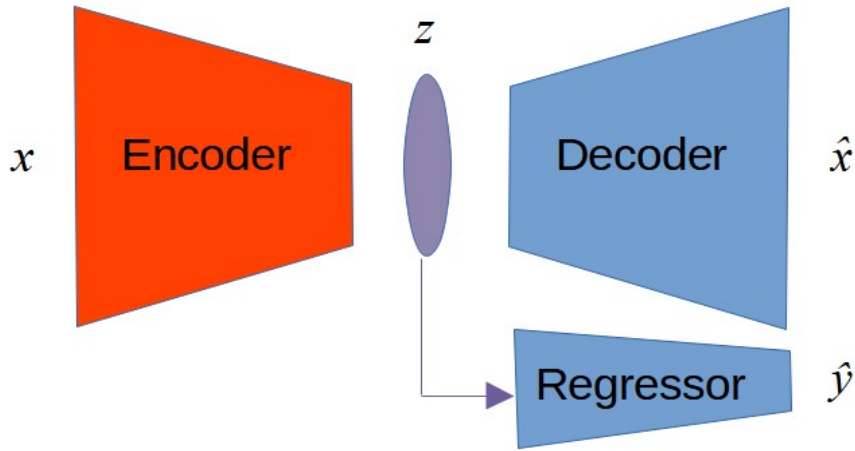

**Figure 2: Schematic of β-VAE based regression model**

In the model discussed above, deterministic dense layers are used with point estimates for the parameter values. These deterministic dense layers are next replaced with probabilistic dense layers in which model parameter priors are represented by normal distribution. Multiple prediction runs over the trained probabilistic layers give estimate of uncertainty. Simplistic assumption of mean-field approximation for latent representation is also replaced with more realistic multi-variate normal distribution for the latent features. Conceptually, this should help in capturing causal relation between factors, and hence be more realistic representation of a complex system (Trauble et al., 2021). Factorized Gaussian prior is replaced with multi-variate normal distribution for the latent features. This approach also helps to directly sample from latent space using joint distribution, and thus avoid the need for reparameterization trick.

# Results

Two DNN models as described in the previous section are trained for both WCF and NPV dataset. In the first model, deterministic hidden layers are used and the latent space is sampled by applying reparameterizaion trick. While in the second model, probabilistic hidden layers are used and latent space is sampled from a joint distribution. WCF dataset has 158 thousand samples while NPV dataset has 126 thousand samples. Due to the discounting applied in NPV calculation, its values show less variance as compared to WCF values. 20% of the samples are used for testing and validation, while the rest are used for training models. Models are trained for 1500 epochs, and ADAM optimizer is used with decreasing learning rate. A scale factor, $\gamma$, of value 25 is used to balance between reconstruction and regression error during training. Quality of test set prediction for two models is analyzed in this section to understand their efficacy in terms of latent representation as well as prediction accuracy and uncertainty quantification. Monte-carlo dropout is used to create a set of 1000 independent predictions for test samples. Prediction mean and standard deviation are calculated using this test output set for the two model. Latent representation is learnt such that both prior Gaussian assumption as well as inductive bias in dataset is incorporated. A meaningful representation would separate the high and low target values as well as show continuous gradation in between regions. This latent representation can then be used in downstream tasks for interpolating target values for unseen samples. As discussed earlier, there are more samples with higher target values as the dataset itself was harvested from separate field optimization studies. Thus, it is expected that for a well-behaved NPV dataset, surrogate models also capture this underlying pattern of relatively low prediction uncertainty for high target values and vice versa.

Table 1 shows the training details, prediction mean squared error and coefficient of determination for two base models, while Figures 3 and 4 show crossplot between prediction mean and standard deviation for the same. These models have a latent dimension size of 90, same as input feature dimension, and beta value of 1 as in original VAE. Both deterministic and probabilistic modeling approach seems to give similar prediction accuracy. WCF case shows higher mean squared error as its samples show higher variance. To help understand the nature of data, the predicted mean values for two cases using the deterministic model are shown in Figure 5. High WCF samples are quite sparse and so trained models show high prediction uncertainty. Next, Figures 6 and 7 show t-SNE projection of the latent embedding along two dimensions for two models. For deterministic model, value gradation is correctly captured but there are gaps in between. On the other hand, probabilistic model is not able to separate high and low values.

**Table 1: Training details and results for two base models**

| Dataset | Model type | Prediction MSE | Prediction $R^2$ score |
|---|---|---|---|
| NPV | Deterministic | 0.36 | 0.945 |
| | Probabilistic | 0.34 | 0.948 |
| WCF | Deterministic | 13.9 | 0.921 |
| | Probabilistic | 14.5 | 0.917 |

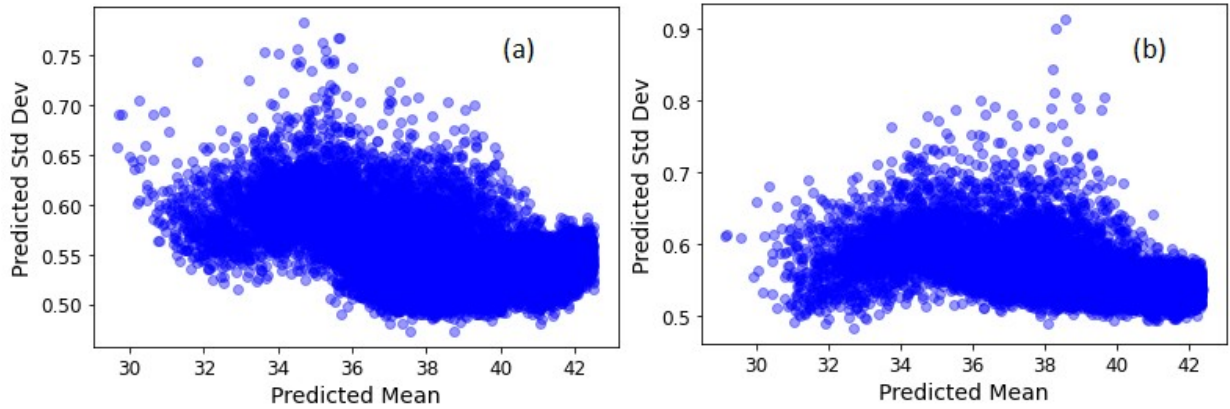
**Figure 3: Crossplot of predicted mean and standard deviation for NPV dataset. (a) Deterministic model, and (b) Probabilistic model.**

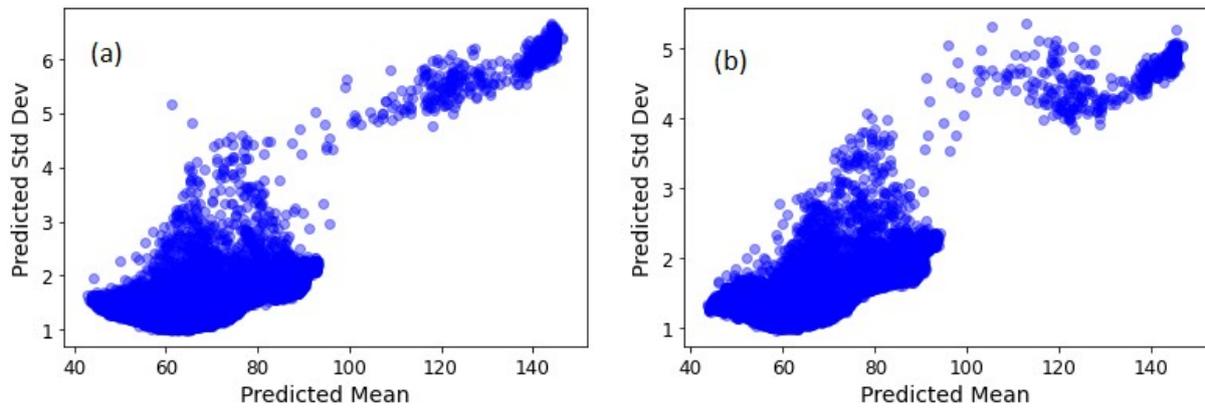
**Figure 4: Crossplot of predicted mean and standard deviation for WCF dataset. (a) Deterministic model, and (b) Probabilistic model.**

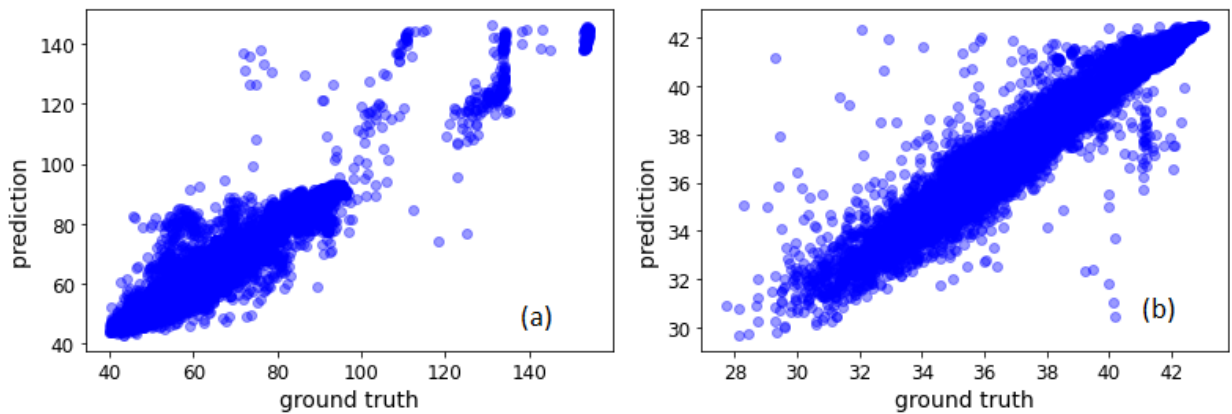
**Figure 5: Crossplot of predicted mean and ground truth for (a) WCF dataset, and (b) NPV dataset**

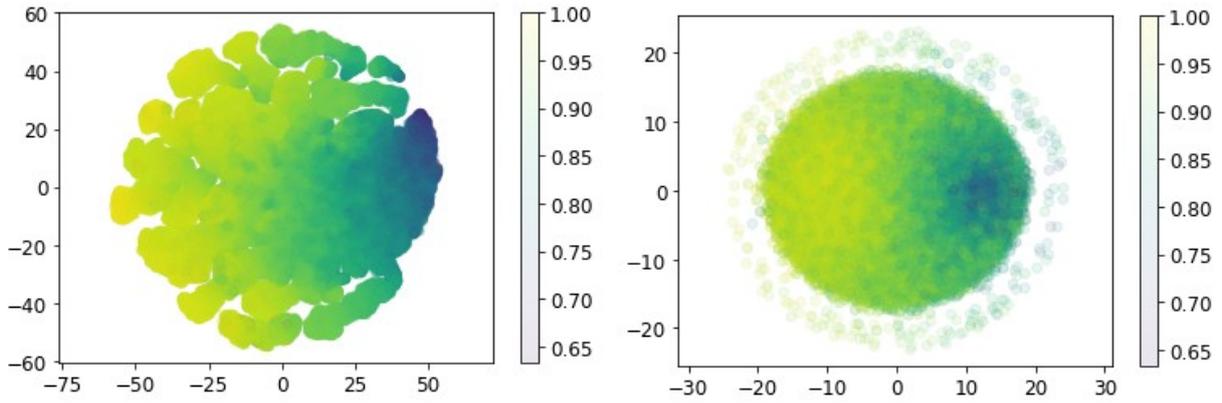

**Figure 6:** t-SNE projection of NPV sample latent embedding along two dimensions for (a) Deterministic model, and (b) Probabilistic model. Point are color coded with scaled NPV values.

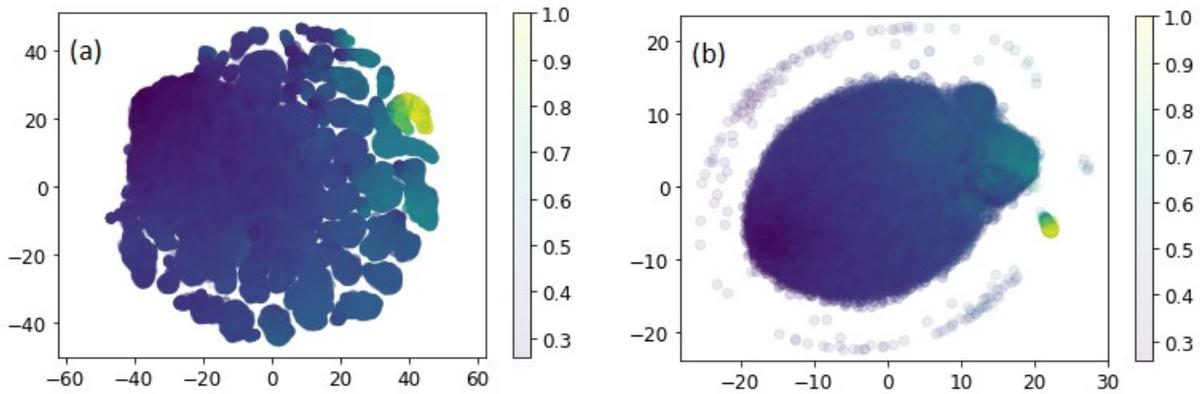

**Figure 7:** t-SNE projection of WCF sample latent embedding along two dimensions for (a) Deterministic model, and (b) Probabilistic model. Point are color coded with scaled WCF values.

Next, we add bottleneck in learning by reducing the latent dimension size to 3, so that models are forced to learn efficient representation. Table 2 shows the training details, prediction mean squared error and coefficient of determination for two models with latent dimension size of 3 and beta value of 1. As can be seen, prediction accuracy remains same even for a lower latent dimension size. Figures 8 and 9 show t-SNE projection of the latent embedding along two dimensions. Probabilistic models are now able to capture the gradation of target values from low to high. Smaller latent dimension size improves packing, and gaps are also reduced.

**Table 2: Training details and results for two models with latent dimension of size 3**

| Dataset | Model type | Prediction MSE | Prediction R2 Score |
|---|---|---|---|
| NPV | Deterministic | 0.35 | 0.947 |
|  | Probabilistic | 0.34 | 0.949 |
| WCF | Deterministic | 14.3 | 0.918 |
|  | Probabilistic | 13.8 | 0.921 |

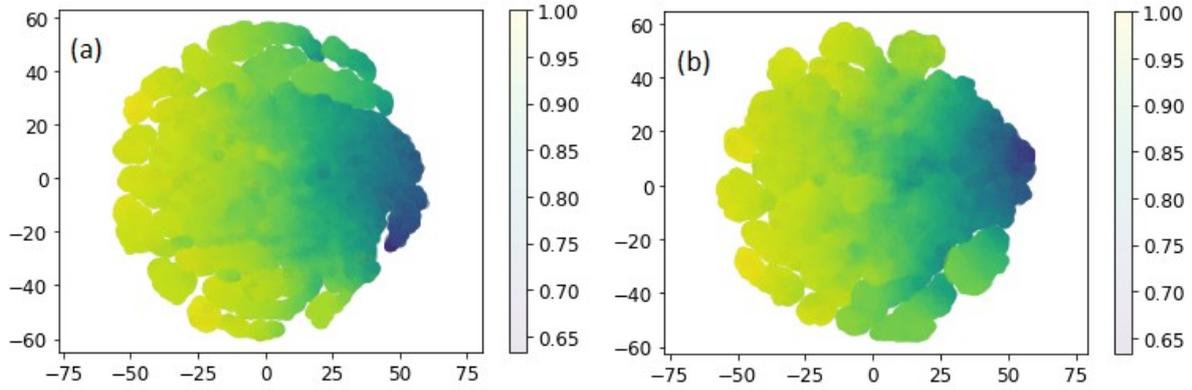

**Figure 8: t-SNE projection of NPV sample latent embedding with latent dimension size of 3 for (a) Deterministic model, and (b) Probabilistic model. Point are color coded with scaled NPV values.**

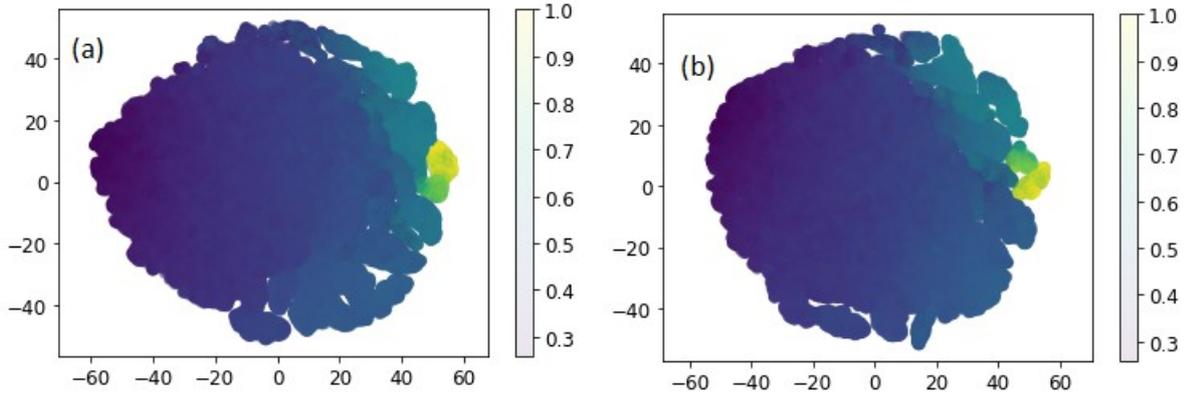

**Figure 9: t-SNE projection of WCF sample latent embedding with latent dimension size of 3 for (a) Deterministic model, and (b) Probabilistic model. Point are color coded with scaled WCF values.**

Finally, we use a beta value of 3 along with latent size of 3 for a more compact representation. Table 3 shows the training details, prediction mean squared error and coefficient of determination for two models with latent dimension size of 3 and beta value of 3. Again prediction accuracy remains similar even for these models. Figures 10 and 11 show t-SNE projection of the latent embedding along two dimensions. Probabilistic models are now able to capture the gradation of target values from low to high as well as reduce gaps in representation, while deterministic models start showing tube like structure in embedding. This is because in deterministic models, extra pressure on KL-divergence term due to larger beta causes posterior to match factorized Gaussian (mean-field) prior. For probabilistic case, this artifact does not show as latent is represented by a multi-normal joint distribution. This is more evident in Figure 12 where beta value is increased to 10 while keeping latent dimension size of 3. While embedding using probabilistic model further improves for higher beta value, deterministic model shows even more prominent tubular artifact.

**Table 3: Training details and results for two models with latent size of 3 and beta value of 3**

| Dataset | Model type | Prediction MSE | Prediction R2 Score |
|---|---|---|---|
| NPV | Deterministic | 0.35 | 0.947 |
|  | Probabilistic | 0.34 | 0.949 |
| WCF | Deterministic | 14.0 | 0.92 |
|  | Probabilistic | 14.0 | 0.92 |

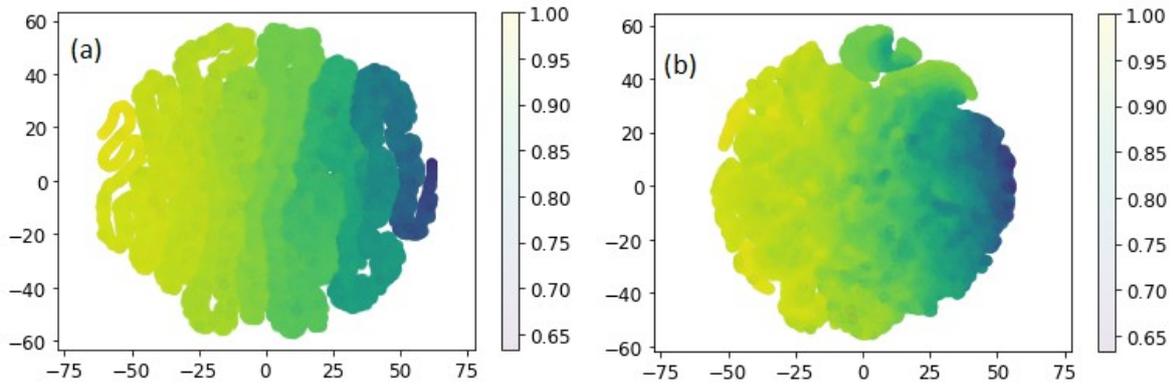

**Figure 10:** t-SNE projection of NPV sample latent embedding with latent dimension size of 3 and beta value of 3 for (a) Deterministic model, and (b) Probabilistic model. Point are color coded with scaled NPV values.

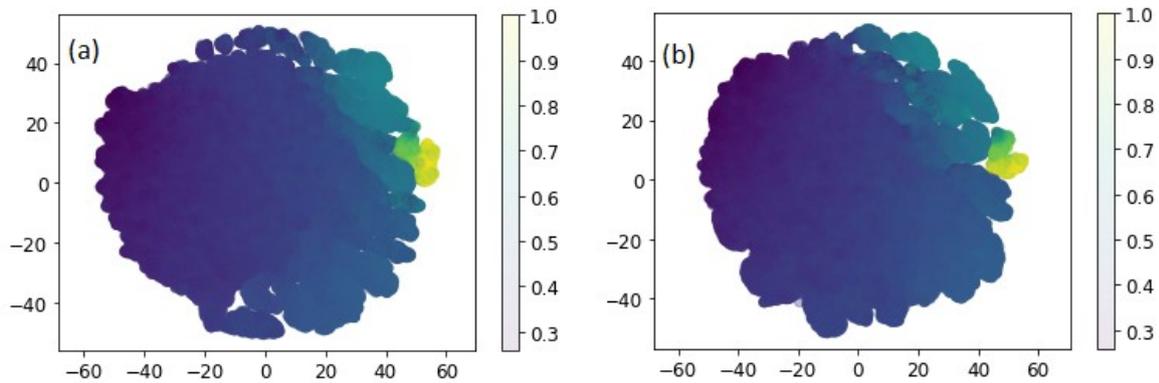

**Figure 11:** t-SNE projection of WCF sample latent embedding with latent dimension size of 3 and beta value of 3 for (a) Deterministic model, and (b) Probabilistic model. Point are color coded with scaled WCF values.

A simplified and meaningful representation can be achieved using β-VAE based regression. By jointly training unsupervised network (β-VAE) and supervised network (regression), inductive bias is injected into representation learning. It can be used to represent high dimensional system in a low dimensional latent embedding. Such a compact representation would be helpful in downstream task like accelerating or enhancing optimization studies. For a field with hundreds of wells, input feature would have relatively large number of dimensions, and a low-dimensional, compact and meaningful representation would be helpful both for other downstream tasks as well as human comprehension.

## Conclusions

Field optimization study of a large oilfield involves many decision variables and are computationally challenging. To accelerate such studies, surrogate of reservoir simulation model is developed using β-VAE based regression. Both VAE and regression network are jointly trained to find a meaningful latent for further use in downstream tasks. Prediction uncertainty is captured both using monte-carlo dropout and probabilistic hidden layers. Multi-variate normal distribution is used in latent space to train correlated factors which are more likely in causal world. Latent dimension size and β parameter is used to find a compact and meaningful representation. Such a representation would be helpful both for other downstream tasks as well as human comprehension.

## Acknowledgments

This research was made possible by generous cloud credits provided by Amazon Web Services, DigitalOcean and Google Cloud. Olympus benchmark reservoir model was made available by the Netherlands Organization for Applied Scientific Research (TNO).

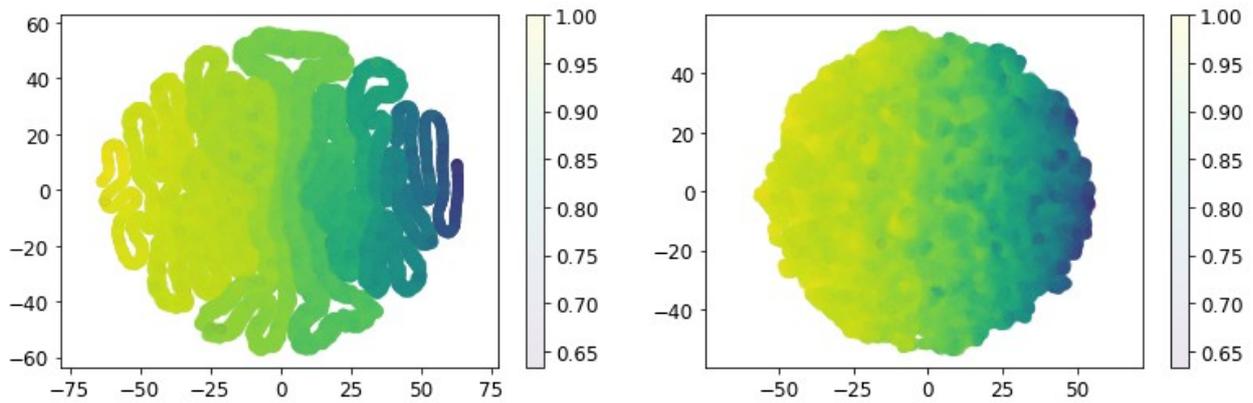

**Figure 12: t-SNE projection of NPV sample latent embedding with latent dimension size of 3 and beta value of 10 for (a) Deterministic model, and (b) Probabilistic model. Point are color coded with scaled NPV values.**

## Nomenclature

| | |
|---|---|
| $\beta$-*VAE* | Beta-variational autoencoder |
| *KL-divergence* | Kullback–Leibler divergence |
| *MCMC* | Markov chain Monte Carlo |
| *MSE* | Mean squared error |
| *NPV* | Net present value |
| $R^2$ | R squared (or coefficient of determination) |
| *t-SNE* | t-distributed stochastic neighbor embedding |
| *WCF* | Weighted sum of cumulative fluid |